\documentclass[a4paper]{article}
\usepackage{hyperref}
\usepackage{INTERSPEECH2020}
\usepackage{array}
\usepackage{lipsum}
\usepackage{multicol}
\usepackage{xcolor}
\usepackage{subfig}
\usepackage{mathtools}
\usepackage{multirow}
\title{End-to-end Named Entity Recognition from English Speech* \newline {\normalsize *submitted to Interspeech-2020}}
\name{Hemant Yadav\textsuperscript{\rm 1}, Sreyan Ghosh\textsuperscript{\rm 1}, Yi Yu\textsuperscript{\rm 2}, Rajiv Ratn Shah\textsuperscript{\rm 1}}

\address{\textsuperscript{\rm 1}MIDAS, IIIT-Delhi, India\\
\textsuperscript{\rm 2}National Institute of Informatics, Tokyo}
\email{hemantya@iiitd.ac.in, gsreyan@gmail.com, yiyu@nii.ac.jp, rajivratn@iiitd.ac.in}

\begin{document}
\maketitle{}
% \listoftables
\begin{abstract}
Named entity recognition (NER) from text has been a widely studied problem and usually extracts semantic information from text. Until now, NER from speech is mostly studied in a two-step pipeline process that includes first applying an automatic speech recognition (ASR) system on an audio sample and then passing the predicted transcript to a NER tagger. In such cases, the error does not propagate from one step to another as both the tasks are not optimized in an end-to-end (E2E) fashion. Recent studies confirm that integrated approaches (\emph{e.g.}, E2E ASR) outperform sequential ones (\emph{e.g.}, phoneme based ASR). In this paper, we introduce a first publicly available NER annotated dataset for English speech and present an E2E approach, which jointly optimizes the ASR and NER tagger components. Experimental results show that the proposed E2E approach outperforms the classical two-step approach. We also discuss how NER from speech can be used to handle out of vocabulary (OOV) words in an ASR system.  

\end{abstract}

\section{Introduction}\label{intro}

Named entities are phrases that contain the names of persons, organizations, locations, others. For example, the sentence,  \emph{T.C.S. CEO Rajesh Gopinathan heads a meeting in their Banglore office}, has the following named entities: [ORG T.C.S.], [PER Rajesh Gopinathan], and [LOC Banglore]. ORG, PER, and LOC represent the organization, person, and location, respectively. In this paper, we focus on these three named entities. 

NER is an important task in information extraction systems and very useful in many applications. It has many progress~\cite{chiu2016named,lample2016neural,nadeau2007survey} and applications such as in optimizing search engine algorithms~\cite{searchengine}, classifying content for news providers~\cite{kumaran2004text} and recommending content~\cite{koperski2017content}. However, despite NER from speech has many applications such as the privacy concerns in medical recordings (\emph{e.g.}, to mute or hide specific words such as patient names)~\cite{cohn2019audio}, it has very limited literature.

% In this paper, we focus on three named entities: persons, locations, and organizations. There has been a lot of work on the extraction of named entities, for textual documents
% % (see Borthwick (1999) for an overview) 
% in the past. One of the use case of NER from speech is privacy concerns in medical recordings, as mentioned by~\cite{cohn2019audio}. The named entity tags information in the speech is be used to mute specific words.

\begin{figure}[ht!]
  \centering
  \includegraphics[width=0.5\linewidth]{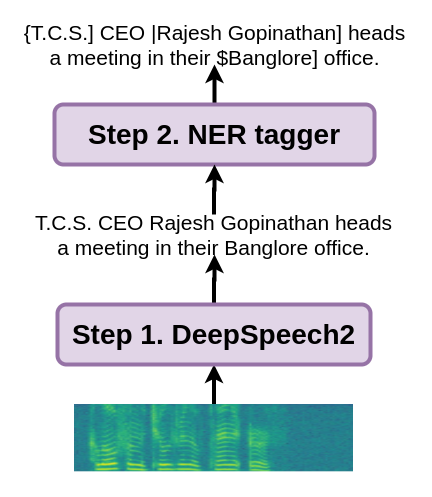}
  \caption{Two-step approach for NER from speech.}
  \label{figure2}
\end{figure}

NER from English speech is done by a classical two-step approach~\cite{cohn2019audio}. It consists of first processing the given audio on an ASR system and then feeding the transcribed ASR output to the NER tagger~\cite{flair2,flair1} (see Figure~\ref{figure2}). Such approaches have several disadvantages, such as existing NER systems are not robust to the noisy output of the ASR, since they are usually designed to process written language. Furthermore, usually, no information corresponding to named entities are used in the ASR system. However, such information could be used to choose better partial hypotheses which are dropped away during the decoding step. As a consequence, even when the decoding goes beyond the 1-best ASR hypothesis for better robustness to ASR errors~\cite{hakkani2006beyond}, the search space is pruned without taking into account knowledge of the partial named entities. In all these cases the NER component is trained independently. Thus, the error does not propagate from one step to another in an E2E fashion.

To the best of our knowledge, the only work related to E2E NER from speech is done on French datasets~\cite{e2ener}. This paper is highly motivated by this work. Additionally, we study the effect of a LM on the E2E NER task. Our major contributions are as follows: (i) we introduce a first publicly available NER annotated dataset\footnote{we will release the dataset for research purpose soon.} for English speech, (ii) we present a state-of-the-art approach for an E2E named entities recognition on the curated dataset, and (iii) we discuss how an E2E system can be used to handle out of vocabulary words in an ASR system.

% Our work is highly motivated by their work. Our major contribution in the paper are: we introduce a NER annotated dataset for English speech, present a baseline study of an E2E approach to extract named entities on the introduced English dataset. Lastly, a discussion on how an E2E system can be used to handle OOV words in an ASR system.
 
The paper is structured as follows. Section~\ref{Related Work} discusses the related work. Section~\ref{dataset} introduces the dataset and Section~\ref{method} describes our methodology. We present our experiments in Section~\ref{experiments}. Section~\ref{discussion} discusses how to deal with the OOV words, and Section~\ref{conclusion} presents the conclusion and future work.

% Named entity recognition (NER) is among spoken language understanding (SLU) tasks to extract semantic information from speech~\cite{tur2011spoken}.

\section{Related work}\label{Related Work}
In the literature many studies~\cite{DBLP:journals/corr/abs-1810-04805,nadeau2007survey,DBLP:journals/corr/abs-1802-05365} focused on NER from textual documents. State-of-the-art (SOTA) NER systems leverage advances in deep learning and recent approaches that take advantage from both word and/or character-level embedding~\cite{baevski2019cloze,jiang-etal-2019-improved,strakova-etal-2019-neural}. However, NER from speech is a less studied problem in the research community. Until very recently, the majority of work in recognizing named entities from speech is done using the two-step approach, including the audio de-identification task~\cite{cohn2019audio}, and the work leveraging OOV information to increase the robustness of NER tagger~\cite{parada2011oov}.

% in their work, for evaluation they have used time-stamps of named entities to calculate the F1 score.  

% Furthermore, ~\cite{sudoh2006discriminative} train an NER model using ASR results with named entity labels to include an ASR confidence feature as well as corresponding transcriptions with NE labels, and from conditional random fields~\cite{hatmi2013named} .

Named entity recognition from speech using an E2E approach has a very limited literature, and moreover there is no work on English speech. Recently, Ghannay et al.~\cite{e2ener} presented an E2E NER on the French datasets. In their work, they used special symbols to achieve NER tagging capabilities in their E2E approach. The special symbols used by them are: “[”, “(”, “\{”,“\$”, “\&”, “\%”, “\#”, “)” and “]” (the first eight symbols to denote the start of 8 different named entities and the last common symbol to denote an end to all of them). Similar to their work, we use three special symbols ('\{','$|$', '\$') to recognize three most frequent named entities (names of organization, person, and location) from our English speech dataset. In this paper, we compare the E2E and the two-step approach for the English speech. Furthermore, we study the effect of a LM on the E2E NER task.
% we clearly compare the results and study both the approaches and show that an E2E approach is better than the 2Step approach.

% \begin{figure*}[ht]
%   \centering
%   \scalebox{0.42}{
%   \includegraphics[]{LaTeX/words.png}}
%   \caption{Example of: True transcript (A), Transcript with the addition of 4 special symbols for training the D-NER model (B) and, Mis-labeled D-NER predicted transcript (C).}
%   \label{figure1}
% \end{figure*}
\newcommand{\RNum}[1]{\lowercase\expandafter{\romannumeral #1\relax}}

\section{Dataset}\label{dataset}

% Information extraction tasks broadly fall into two categories: subjective and objective tasks~\cite{liu2010sentiment}. Subjectivity refer to attitudes, opinions, and emotions. They are subjective impressions and include the work on: sentiment analysis~\cite{sent1,sent2}, text categorization~\cite{text_cate1, text_cate2} and emphasis detection~\cite{emphasis}. Objective categories include facts like names and location.  

The annotated English dataset we prepare for the NER from speech task is a subset of a combination of Librispeech~\cite{librispeech}, CommonVoice~\cite{commonvoice}, Tedlium~\cite{tedlium} and Voxforge~\cite{voxforge}. The recordings are comprised mainly of two domains: Reading English and Ted talks. We refer to this combined data as DATA1, which has around 600,000 files (approximately 1,000 hrs). After an empirical analysis, we found that the majority of files did not have any named entities. To remove these files from the manual annotation step, we use Flair~\cite{flair1,flair2} as a NER tagger with 0.9 F1 score. In this way, we have reduced the number of files having NER to 70,769, which are used for manual annotation.

\begin{figure}[ht]
\centering
    \scalebox{1}{\begin{tabular}{|p{7cm}|}
    % \hline
        %  \textbf{} & \textbf{Transcript} \\
    \hline
         \textbf{Character sequence:} T.C.S. CEO Rajesh Gopinathan heads a meeting in their Banglore office.\\
    % \hline
        \textbf{Character sequence with named entities:} [ORG T.C.S.] CEO [PER Rajesh Gopinathan] heads a meeting in their [LOC Banglore] office.\\
    % \hline
         \textbf{Character sequence with special symbols:} \{T.C.S.] CEO $|$Rajesh Gopinathan] heads a meeting
          in their \$Banglore] office.\\
    % \hline
        % \textbf{Mis-labeled character sequence:} \hspace{0.1cm}T.C.S. \{CEO] $|$Rajesh Gopinathan] heads a meeting in their \$Banglore] office.\\
    % \hline
    \hline
    \end{tabular}}
    \caption{Example of mapping the named entities. \textbf{'$|$', '\$', '\{'} denote the start of a named entity and \textbf{'$]$'} denotes the end.}
\label{figure1}

\end{figure}

Thus, the dataset is prepared into two steps: (\RNum{1}) applying a NER tagger on DATA1 (600,000 files) and (\RNum{2}) manually annotating the 70,769 files having valid NERs using Doccano\footnote{https://github.com/doccano}~\cite{doccano}. 
In step 1, we re-train the Flair tagger on the capitalized NER benchmark CoNLL-2003~\cite{sang2003introduction} dataset from scratch. The transcripts at the test time are capitalized, and so is the training data. Furthermore, after an empirical analysis, it was found that a threshold probability of 0.95 rejects the majority of noisy/erroneous named entities from the tagger output. After the Flair NER tagger operation on DATA1, the total number of files is reduced to 70,769 (approx. 150 hrs) from 600,000 (approx. 1,000 hrs). We refer to this reduced data as DATA2. Since DATA1 has audios of different speakers recording the same sentence, DATA2 also has repetitions. To be precise, DATA2 has a total of 31,000 unique sentences, and the rest 39,769 are a repetition of 1238 sentences from 31,000 unique sentences. 

In step 2, We manually annotate all the remaining 70,769 files in DATA2 following CoNLL-2003~\cite{sang2003introduction} guidelines. An example of the manually annotated sentence is shown in Figure~\ref{figure1}, i.e., character sequence with and without the named entities. To increase the robustness of the model, similar to~\cite{6685834}, we asked the annotator to randomly mislabel some tokens as named entities (\emph{e.g.}, annotating the CEO token as [PER] or Banglore token as [ORG]).

\begin{table}[ht!]
\setlength{\tabcolsep}{5pt}
\caption{Category wise distribution of unique and repeated sentences in DATA2.}
\label{table:data}
\centering
    \scalebox{1}{\begin{tabular}{cccc}
    \hline
         \textbf{Category} & \textbf{Unique} & \textbf{Repetition} & \textbf{Total}\\
    \hline
         Person & 24711 & 20268 & 44979\\
    % \hline
         Location & 11881 & 7948 & 19829\\
    % \hline
         Organization & 2299 & 1473 & 3772\\
    \hline
        Total & 38891 & 29689 & 68580\\
    \hline
    \end{tabular}}
\end{table}

The named entities distribution in DATA2 are shown in Table~\ref{table:data}, and DATA2 has a total of 38891 unique named entity tokens, as shown in the last row of the Table~\ref{table:data}. Furthermore, DATA2 is comprised of 34\% Librispeech, 36\% CommonVoice, 7\% Tedlium, and 23\% Voxforge of DATA1.
% The reader should have in mind that DATA2 is used for all the experiments in this study, and the data has a class imbalance. 

% for D-NER model. 

% The annotated sentences are saved in default dictionary format of Doccano~\cite{doccano}.

% , with keys being 'text' and 'labels'. Labels have the named entity and their start and end position.
% \begin{figure}
%     \centering
%     \includegraphics[width=0.25\textwidth]{LaTeX/annotaion.png}
%     \caption{Annotation steps to create the dataset.}
%     \label{fig:annotation}
% \end{figure}

\section{Methodology}\label{method}

This section provides information on the E2E approach used in this study, the training method used to train it and the different components used in the decoding step at the test time.

\subsection{Language Model}\label{language model}

A language model (LM) is a probability distribution over an arbitrary symbol sequences P(w\textsubscript{1}, ..., w\textsubscript{n}) such that more likely sequences are assigned higher probabilities and vice versa. LMs are frequently used at the decoding step to condition beam search. During the decoding, the top-n candidates are evaluated by conditioning the output of acoustic model with the language model. In this study, 4-gram LM is used and is trained on the full dataset i.e., the combined train, dev, and test using the KENLM library~\cite{kenlm}.

% In all the experiments make use of a 4-gram language model and "prefix beam search"~\cite{prefixbeamsearch} decoding.  The hyper-parameters for the "Prefix beam search" decoding~\cite{prefixbeamsearch} are: beam-width of 1024 and 1.96 \& 6.0 in place of alpha \& beta values.

\subsection{E2E approach}\label{methodology}
The problem of NER from speech is to assign a special symbol before and after a named entity to identify it. We approach this problem as a sequence labeling task and use an RNN based Baidu's DeepSpeech2 (DS2)~\cite{deepspeech2} neural architecture to study NER from speech with modifications in the last layer. DS2 is a combination of Convolution Neural Network (CNN) and Recurrent Neural Network (RNN) layers, with a fully connected followed by a softmax layer. The softmax layer outputs the probabilities of the sequence of characters.

% and a $<$BLANK$>$ symbol (e.g., LLLEEETTTT$<$BLANK$>$TTEEERRRR). After collapsing the characters (e.g., LET$<$BLANK$>$TER), the $<$BLANK$>$ symbols are removed (e.g., LETTER)from the sequence of output characters to get the final predicted transcript. 

\begin{figure}[ht!]
  \centering
  \includegraphics[width=0.95\linewidth]{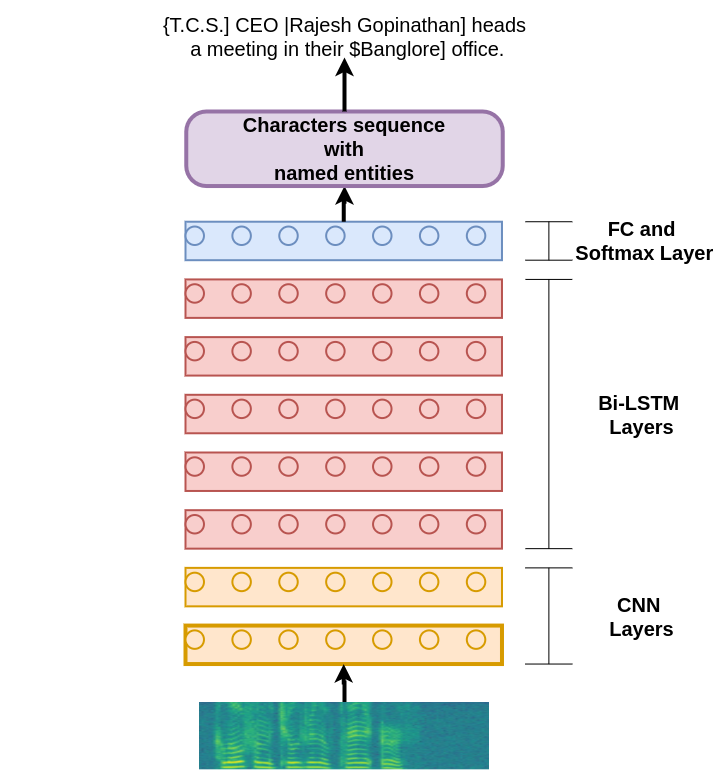}
  \caption{E2E approach for the NER task.}
  \label{figuree2e}
\end{figure}

Let $X = \{x_1, x_2, ..., x_n\}$ be the input utterances and $Y = \{y_1, y_2, ..., y_n\}$ be the corresponding transcripts. For a given input audio $x_1$, it is transformed into a sequence of "log-spectrograms of power normalized audio clips, calculated on a 20ms window", and is then fed to the model, which captures the sequential nature of speech. The output $l$ is a sequence of defined set of characters ('A-Z' + ' '). Similar to the work~\cite{e2ener}, we add special symbols ('$|$', '\$', '\{' denote the start of named entities person, location and organization, respectively, and '$]$' denotes the end of any three named entities) to the pre-defined set of characters to introduce the named entity tagging capabilities in the architecture. An example of a character sequence with and without the special symbols is shown in the Figure~\ref{figure1}. To recognize named entities, we modify the DS2 architecture by increasing the shape of the fully connected layer by four to accommodate the extra symbols in the output layer (see Figure~\ref{figuree2e}).

\begin{figure}[ht!]
  \centering
  \includegraphics[width=0.28\textwidth]{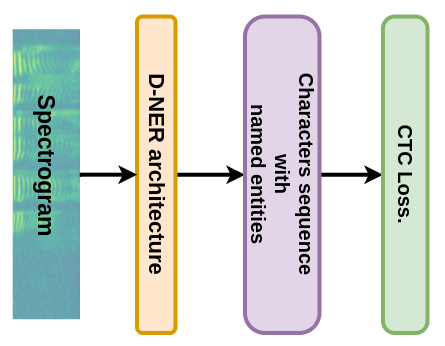}
  \caption{Training of the E2E architecture.}
  \label{figuretrain}
\end{figure}

We train the system using the Connectionist temporal classification~\cite{ctc} (CTC) loss (see Figure~\ref{figuretrain}) because CTC loss takes into account all the possible character sequences given the output and the true transcript. Thus, CTC loss maximizes the total probability of all the paths which lead to the true transcript. The model predicts $p(l\textsubscript{t}/x_1)$ at each time step.

At the test time, the output is conditioned on an N-gram LM using prefix beam search decoding~\cite{prefixbeamsearch}, is shown as follows.

\[Q(y) = log(p(l\textsubscript{t}/x)) + \alpha*log(pLM(y)) + \beta*wc(y) \]

Where $wc(y)$ is the word count in the predicted transcript. $\alpha$ and $\beta$ control the contribution of LM and the number of words in the predicted transcript. In this study, we use values 1.96 \& 6.0 for $\alpha$ and $\beta$, respectively.

To compensate for the limited data for the NER from speech task, DS2 weights are used as a starting point to train the E2E model. We train a standard DS2 architecture on the DATA1 as a baseline. This system achieves a word error rate (WER) of 2.72\% on the test set with a 4-gram LM and beam-width equal to 1024. In a similar work on the French dataset~\cite{e2ener}, they achieved a WER of 19.96\%. Better WER is the major reason that our E2E NER system achieves better results compared to~\cite{e2ener}.  

% The reason mentioned in their work was the size of the dataset available for training. Better WER is the major reason that our E2E system achieves far better results compared to their work.

% \begin{table}[ht]
% \setlength{\tabcolsep}{10pt}
% \renewcommand{\arraystretch}{1}
%     \centering
%     \caption{Named Entities distribution: Train, Dev and Test. }
%     \label{table:tdt}
%     \begin{tabular}{cccc}
%     \hline
%          Category & Train & Dev & Test\\
%     \hline
%          Person & 19721 & 2478 & 2512\\
%     \hline
%          Location & 9416 & 1250 & 1215\\
%     \hline
%          Organization & 1832 & 241 & 226\\
%     \hline
%         Sum & 30960 & 3969 & 3953\\
%     \hline
%     \end{tabular}

% \end{table}

\section{Experiments}\label{experiments}

In this section we discuss the experimental setup of E2E and two-step approaches for NER from speech, the method used for evaluating the two systems and the corresponding results.

\subsection{Experimental Setup}\label{experimental setup}
All the experiments were carried out on DATA2 prepared in Section~\ref{dataset}. Both the dev and test set are created from the 31,000 unique files with a 10\% distribution each and the remaining files in DATA2 are used for training. 

% Additionally, the train, dev and test distribution of each named entity is around 80\%, 10\%, 10\% respectively.

We experimented with two different approaches for the NER from speech task: E2E and a classical two-step approach. For the E2E approach, experiments were carried out on the model explained in Section~\ref{methodology}. At the test time, Prefix beam search decoding with a beam-width of 1024 and a 4-gram LM (trained on DATA2) is used. In the case of classical two-step approach, we use Baidu's DS2 as the ASR component and Flair as a NER tagger component. In the two-step approach, audio is first transcribed using the ASR component, and then the output is passed to the Flair tagger as shown in the Figure~\ref{figure2}. The Flair tagger used in the two-step approach is explained in the Section~\ref{ner tagger}.

\subsection{NER Tagger}\label{ner tagger}
NER, also known as entity identification, and entity extraction, seeks to locate and classify named entities in text into some pre-defined labels. For this study, we use Flair as the NER tagger in the two-step approach. The Flair tagger is trained on a combined dataset of DATA2 and CoNLL-2003~\cite{sang2003introduction} and the test set is same as in the case of the E2E approach. The results for the Flair NER tagger on the test set are shown in Table~\ref{table:2step tagger}.  

\begin{table}[ht!]
\centering
    \caption{Precision, Recall and F1 score of the Flair NER tagger used in place of the step 2 in classical two-step approach.}
    \label{table:2step tagger}
    \begin{tabular}{cccc}
    \hline
         \textbf{Category}  & \textbf{Precision} & \textbf{Recall} & \textbf{F1}\\
    \hline
         Person & 0.84 & 0.88 & 0.86\\
    % \hline
         Location & 0.87 & 0.86 & 0.87\\
    % \hline
         Organization & 0.86 & 0.77 & 0.81\\
    \hline
    \end{tabular}

\end{table}

\subsection{Evaluation}\label{evaluation}
Similar to the work~\cite{e2ener}, we use F1 score~\cite{sasaki2007truth} for evaluation, which is defined as follows.

\[F_{1}=\frac{2\times\text{Precision}\times\text{Recall}}{ \text{Precision} + \text{Recall}}\]

Precision is the percentage of named entities that are correct in the predictions compared to true labels and recall is the percentage of ground truth named entities found by the system. A named entity is correct only if it is an exact match to the ground truth entity. We emphasize on using micro average since there is a class imbalance as shown in Table~\ref{table:data}. 

To consider the case of repeated tags and half-labeled predictions, we slightly modify the precision and recall calculation. For example, if a sentence has two identical named entities, then we collapse them and treat them as one. Secondly, we discard any tags which are half-labeled (e.g., in \{T.C.S.] CEO $|$Rajesh Gopinathan heads a meeting in their \$Banglore] office. There is no end label to the PER tag.). Apart from these two special cases, we followed all the standard guidelines for calculating precision and recall.

% \begin{table}[ht]
% \setlength{\tabcolsep}{10pt}
% \renewcommand{\arraystretch}{1}
% \centering
% \caption{WER: standard DS2 and Modified DS2, with and without LM.}
% \label{table:1}
%     \begin{tabular}{ccc}
%     \hline
%          system & wer & LM\\
%     \hline
%          Standard DS2 & 19.361 & no   \\
%     \hline
%          Standard DS2 & 2.715 &  yes\\
%     \hline
%          Modified DS2 & 25.168 & no\\
%     \hline
%          Modified DS2 & 4.969 & yes\\
%     \hline
%     \end{tabular}
% \end{table}

\subsection{Experimental Results}\label{results}

In this section, we report results on the DATA2 dataset. In all the experiments, we use batch normalization and weight decay as regularization. The evaluation metrics we report are precision, recall, and F1 score on the NER task. The reader should have in mind that the half-labeled tags are discarded for all the calculations, as mentioned in the Section~\ref{evaluation}. 
% Additionally, the data has a class imbalance, as shown in Table~\ref{table:data}, and therefore we recommend and prefer micro scores over macro. 
Pre-trained models and training configurations are available on GitHub\footnote{link to the GitHub repository.}.

\begin{table}[ht ]
\centering
\caption{Precision, Recall and F1 score of the two-step and E2E NER from speech, with the LM.}
\label{table:NER results}
    \scalebox{0.95}{
    \begin{tabular}{ccccc}
    \hline
         \textbf{System} & \textbf{Category}  & \textbf{Precision} & \textbf{Recall} & \textbf{F1}  \\
    \hline
          \multirow{5}{*}{Two-step} & Person & 0.82 & 0.82 & 0.82 \\ 
                  & Location & 0.83 & 0.79 & 0.81 \\
                  & Organization & 0.75 &   0.16 & 0.27 \\
                  & Micro average & 0.83 &  0.77 & 0.80 \\
                  & macro average & 0.80 &  0.59 & 0.63 \\
    \hline
     
         \multirow{5}{*}{E2E NER}& Person & \bf0.96 & \bf0.86 & \bf0.91 \\
                  & Location & \bf0.97 &  \bf0.85 & \bf0.91 \\
                  & Organization & \bf0.96 &  \bf0.70 & \bf0.81 \\
                  & Micro average & \bf0.96 &  \bf0.85 & \bf0.90 \\
                  & macro average & \bf0.96 &  \bf0.80 & \bf0.87 \\
    \hline
        \end{tabular}}

\end{table}

Table~\ref{table:NER results} shows our experimental results on the E2E and two-step approach. The scores are in order of person, location, and organization from better to worse because of the class imbalance present in the DATA2, as shown in Table~\ref{table:data}. Furthermore, our results prove that the E2E approach outperforms the classical two-step approach in all the cases and by a significant margin. Moreover, the recall for both the approaches is lower compared to the precision. we think that it is because of the smaller size of dataset for the task. The similar trend of precision and recall is observed in the work on French datasets~\cite{e2ener}.

% and achieved 0.47 F1 score compared to our 0.90.  

We also studied the effect of language model on the F1 scores. LM improves the NER scores with a significant margin, and the same can be inferred from Table~\ref{table:with and without lm}, which shows an improvement of almost by a factor of 300\% with LM. Therefore, based on the quantitative analysis, we conclude that the NER results are closely dependent on the language model, and if an LM is trained on a bigger corpus, then the recall could be increased further.

Lastly, if we look at the Table~\ref{table:2step tagger}, the F1 scores for detecting a named entity is less compared to the E2E approach as shown in Table~\ref{table:NER results}. Therefore, even if we can get the perfect transcriptions from the ASR component, the best we could do is still less than the E2E approach. The reason could be, in the E2E approach the output and the LM (trained with the named entity tags) are conditioned together in the decoding step. Thus, two sources of information are taken into consideration in the E2E approach. Further analysis is required to make any concrete comments.

\begin{table}[ht!]
\centering
\caption{E2E NER from speech: Micro-Average scores, with and without LM. }
\label{table:with and without lm}
    \begin{tabular}{cccc}
    \hline
         \textbf{E2E NER}  & \textbf{Precision} & \textbf{Recall} & \textbf{F1}\\
    \hline
         without LM & 0.38 & 00.21 & 0.27\\
    % \hline
         with LM & \bf0.96 &  \bf0.85 & \bf0.90 \\
    \hline
    \end{tabular}

\end{table}

% Therefore, based on our experiments, we now conclude that the E2E is better than the two-step approach and outperform it by a significant margin.

\section{Handling OOV words in an ASR system}\label{discussion}
In this section, we discuss how the information on named entities can be used to handle OOV words in an ASR system.

Let us start with some statistics concerning OOV words. We trained the LM on the training data only and found out that the percentage of OOV words in development and test set is around 20\% each. Out of those 20\% OOV words, around 40\% are named entities. If an LM is trained on a significantly bigger corpus, than the share of named entities in the OOV words will increase. Thus most of the OOV words will be the named entities. From this point on, the discussion will be based on just one named entity i.e., person.

\begin{figure}[ht!]
        \scalebox{1}{
        \begin{tabular}{p{4cm}p{4cm}}
        % \hline
            My name is \textcolor{red}{Gaurav Yadav}. & \\
            My name is \textcolor{red}{Modi}. & My name is \textcolor{red}{$<$PER$>$}. \\
            My name is \textcolor{red}{Rajiv}. & \\
        % \hline
        \\
        \textbf{(a) Standard sentences.} & (b) \textbf{Modified sentence.} \\
        \end{tabular}}

\caption{Sentences to train the LM.}
\label{figure3}
\end{figure}

% An LM has the probability of all the tokens arranged in a specific manner. 
So far, from the E2E model, we have the probability of a token being a named entity, and only if we could have the same information from the LM, we can condition it using the modified prefix beam search decoding. This can be achieved, if we replace all the individual person tokens with a $<$person$>$ token as shown in Figure~\ref{figure3} and train an LM (we call it semantic LM or S-LM) on the updated corpus again. The S-LM would learn the probability of the next token being a $<$PER$>$ instead of the exact names. Therefore, at the test time, we now have the named entity tag information from the E2E model and the S-LM. At the test time, now we can condition the E2E model output with the S-LM to rank the top n-paths using the modified prefix beam search. The modified "prefix beam search" decoding is such that, it assigns a higher probability to the output of E2E model if the same can be inferred from the S-LM or otherwise. Furthermore, while scoring the top beams, we can condition the named entities on actual person names from a pre-defined dictionary of names. 

Since we do not have a concept of an individual word now but a $<$PER$>$ token. Therefore, the problem of named entities being an OOV word is not anymore. Thus we can now condition the E2E model output heavily on the LM, worrying less about penalizing the OOV words.

\section{Conclusion and Future Work}\label{conclusion}
In this paper, we made available a first public English speech dataset with named entities. We also presented a detailed comparison between the E2E and the two-step approaches for NER from speech. Experimental results show that the E2E approach provides better results (F1=\textbf{0.906}) compared to the two-step approach (F1=0.803). Additionally, a LM plays an important role to achieve these numbers. It is the first study to recognize named entities in English speech using an E2E approach. To conclude, this study presents promising results in a first attempt to experiment with the E2E approach to recognize named entities and constitutes an interesting start point for future work. In the future, we can study the effect of other loss metrics including: NE-WER (Named Entity Word Error Rate)~\cite{newer} and ATENE (Automatic Transcription Evaluation for Named Entity)~\cite{atene}, instead of just using WER. Additionally, we can further work on our discussions on handling OOV words in an ASR system.

% and introduce a modified "prefix beam search" decoding which takes into consideration the information of named entity tags to calculate the top-n best hypothesis.

\bibliographystyle{IEEEtran}
\bibliography{mybib}

\end{document}